\documentclass[sigconf]{acmart}

\usepackage[dvipsnames]{xcolor}
\usepackage{pifont}
\usepackage{xcolor}
\usepackage{enumitem}
\usepackage{tabularx}
\usepackage{booktabs}
\usepackage{caption}
\usepackage{xspace}
\makeatletter
\let\oldfootnote\footnote
\def\footnote{\@ifstar\footnote@star\footnote@nostar}
\def\footnote@star#1{{\let\thefootnote\relax\footnotetext{#1}}}
\def\footnote@nostar{\oldfootnote}
\makeatother

\newcommand{\cmark}{\textcolor{green!70!black}{\ding{51}}}
\newcommand{\xmark}{\textcolor{red!70!black}{\ding{55}}}

\newcommand{\ourmethod}{\textsc{EpiCF-Bench}\xspace}
\AtBeginDocument{%
  \providecommand\BibTeX{{%
    \normalfont B\kern-0.5em{\scshape i\kern-0.25em b}\kern-0.8em\TeX}}}

\copyrightyear{2026}
\acmYear{2026}
\setcopyright{cc}
\setcctype{by-nc-nd}
\acmConference[KDD '26]{Proceedings of the 32nd ACM SIGKDD Conference on Knowledge Discovery and Data Mining V.2}{August 09--13, 2026}{Jeju Island, Republic of Korea}
\acmBooktitle{Proceedings of the 32nd ACM SIGKDD Conference on Knowledge Discovery and Data Mining V.2 (KDD '26), August 09--13, 2026, Jeju Island, Republic of Korea}
\acmDOI{10.1145/3770855.3817522}
\acmISBN{979-8-4007-2259-2/2026/08}

\begin{document}

\title{Dynamic Counterfactual Benchmark via Differentiable Agent-Based Model}
\title{A Benchmark for Counterfactual Prediction in Epidemic \\Time Series with Dynamic Interventions}
\title{Benchmarking Counterfactual Prediction in Epidemic Time Series with Time-Varying Interventions}
%


\author{Wenhao Mu}
\authornote{Both authors contributed equally to this research.}
\email{muwenhao@umich.edu}
\affiliation{%
  \institution{University of Michigan}
  \department{Computer Science and Engineering}
  \city{Ann Arbor}
  \state{Michigan}
  \country{USA}
}

\author{Facundo Yan}
\authornotemark[1]
\email{facundoy@umich.edu}
\affiliation{%
  \institution{University of Michigan}
  \department{Computer Science and Engineering}
  \city{Ann Arbor}
  \state{Michigan}
  \country{USA}
}

\author{Anik Mumssen}
\email{mumssen@umich.edu}
\affiliation{%
  \institution{University of Michigan}
  \department{Computer Science and Engineering}
  \city{Ann Arbor}
  \state{Michigan}
  \country{USA}
}

\author{Marisa Eisenberg}
\email{marisae@umich.edu}
\affiliation{%
  \institution{University of Michigan}
  \department{Epidemiology \& Complex Systems}
  \city{Ann Arbor}
  \state{Michigan}
  \country{USA}
}

\author{Alexander Rodríguez}
\email{alrodri@umich.edu}
\affiliation{%
  \institution{University of Michigan}
  \department{Computer Science and Engineering}
  \city{Ann Arbor}
  \state{Michigan}
  \country{USA}
}


\renewcommand{\shortauthors}{Wenhao Mu, Facundo Yan, Anik Mumssen, Marisa Eisenberg, \& Alexander Rodríguez}

\begin{abstract}
Deep learning has enabled significant advances in time-series causal inference, yet progress remains constrained by the lack of realistic benchmarks with observable counterfactual outcomes. Existing datasets either rely on real-world observations without ground-truth counterfactuals or on simplified simulations that fail to capture complex causal dynamics. To address this gap, we develop a large-scale benchmark for counterfactual prediction in epidemic time series under dynamic interventions. Unlike existing benchmarks, it supports static and time-varying treatments, as well as both single-policy and multi-policy intervention settings, enabling evaluation of causal inference methods across a broad range of causal inference scenarios. Leveraging a calibrated agent-based model grounded in real-world demographic, mobility, epidemiological, and policy data, we generate realistic counterfactual trajectories across more than 150 U.S. counties. Using this benchmark, we evaluate widely used and state-of-the-art causal inference methods, revealing substantial performance differences and highlighting the challenges of realistic time-series causal reasoning. 
\end{abstract}




\maketitle
\newcommand\kddavailabilityurl{https://doi.org/10.1145/3770855.3817522}
\ifdefempty{\kddavailabilityurl}{}{
\begingroup\small\noindent\raggedright\textbf{Resource Availability:}\\
The source code of this paper has been made publicly available at \url{https://github.com/complex-ai-lab/epi-cf-benchmark}.
\endgroup
}

\section{Introduction}\label{sec:intro}
Understanding the effects of interventions is central to decision-making in domains such as epidemiology, economics, and public policy. In many of these settings, outcomes and interventions evolve over time, making time series data a natural framework for causal inference~\cite{granger1969investigating, moodie2007demystifying}. Unlike static settings, dynamic systems exhibit temporal dependencies, delayed treatment effects, and feedback between interventions, covariates, and outcomes. For example, epidemic control policies may influence disease transmission only after several days or weeks, while behavioral factors such as population mobility evolve alongside the epidemic and affect future policy decisions. These characteristics make time-series causal inference substantially more challenging than its static counterpart.

Recent advances in machine learning have produced a wide range of methods for causal inference in time series, from meta-learning approaches to modern generative models~\cite{wu2024counterfactual,mu2025counterfactual}. Yet progress remains constrained by the lack of realistic benchmark datasets. Real-world datasets, such as MIMIC-III~\cite{johnson2016mimic}, provide rich temporal dynamics but lack observable counterfactual outcomes, making it impossible to directly evaluate counterfactual predictions and treatment effects, and often suffer from confounding, making rigorous evaluation difficult~\cite{moraffah2021causal,pearl2009causal}. In contrast, synthetic and semi-synthetic benchmarks, including~\cite{ferdous2025timegraph}, provide ground-truth counterfactuals but are typically generated from simplified simulations that do not faithfully capture real-world dynamics. This creates a fundamental trade-off between realism and evaluability, motivating the need for benchmarks that bridge both worlds.

To address this gap, we introduce \ourmethod, a benchmark suite for counterfactual prediction in epidemic time series under time-varying interventions. Built upon a calibrated agent-based model (ABM), it generates factual\footnote{``Factual’’ refers to the epidemic trajectory simulated by the ABM when conditioned on the observed sequence of real-world policy interventions.} and counterfactual epidemic trajectories using real-world demographic, mobility, epidemiological, and intervention data. Spanning more than 150 U.S. counties, the benchmark provides large-scale time-series data with diverse treatments, outcomes, and covariates, enabling rigorous evaluation of causal inference methods under dynamic interventions.

Using this benchmark, we establish evaluation tasks covering both single-policy and multi-policy intervention settings and compare a range of widely used and state-of-the-art causal inference methods. We summarize our contributions as follows:
\\
\noindent $\bullet$ \textbf{EpiCF-Bench: Bridging realism and evaluability in time-series causal inference.}
\textsc{EpiCF-Bench} is a large-scale benchmark for causal inference in time series built from real-world demographic, mobility, epidemiological, and intervention data. It provides factual and counterfactual trajectories across 158 U.S. counties, enabling rigorous evaluation of counterfactual prediction and treatment effect estimation under dynamic intervention settings.
\\
\noindent $\bullet$ \textbf{Scalable and policy-aware counterfactual simulation.}
We build on differentiable and tensorized ABM frameworks to construct a scalable epidemic simulation pipeline tailored for counterfactual time series generation. Our framework introduces mechanisms that improve realism and causal relevance, including underreporting correction, time-varying mobility networks, policy characterization from real-world intervention data, and policy implementation through network-level modifications. 
\\
\noindent $\bullet$ \textbf{Comprehensive benchmark suite and open-source resources.}
Leveraging \textsc{EpiCF-Bench}, we establish benchmark tasks covering both single- and multi-policy time-varying intervention settings and evaluate widely used and state-of-the-art causal inference methods. We also release the benchmark datasets, simulation framework, calibration methodology, and data-generation code to support future research on causal inference in dynamical systems.

\section{Related Work}\label{sec:related}
\subsection{Agent-based Models (ABMs)}
Agent-based models (ABMs) are computational simulations in which individual agents interact according to predefined behavioral and environmental rules~\cite{bonabeau2002agent,holland1991artificial,rodriguez2024machine}. Owing to their ability to represent heterogeneous populations and complex interactions, ABMs have been widely applied across domains ranging from socioeconomic policy analysis~\cite{holland1991artificial,zheng2022ai} to biological and medical systems~\cite{ruiz2022simulations,glen2019agent}. In epidemiology, ABMs provide a natural framework for modeling disease transmission and evaluating intervention strategies at the population level~\cite{marathe2013computational,abueg2021modeling}. Prior work has developed increasingly realistic epidemic simulators that capture individual-level disease progression, contact-network dynamics, and mobility-driven interactions~\cite{hinch2021openabm,aylett2021june}. To support large-scale simulations, researchers have also proposed efficient implementations and distributed computing strategies that substantially reduce the computational cost of epidemic modeling~\cite{eames2015six,pellis2015eight,bisset2009epifast}.

More recently, \citet{chopra2023differentiable} proposed GradABM, a differentiable ABM framework for end-to-end gradient-based calibration of epidemic simulators. By representing agent states as tensors, interaction networks as sparse matrices, and non-differentiable operations through differentiable approximations, GradABM enables efficient large-scale simulation and gradient-based parameter estimation. This significantly reduces the computational burden of calibration and enables simulators to be directly fitted to real-world observations. While prior work has demonstrated the utility of differentiable ABMs for epidemic forecasting and policy evaluation~\cite{chopra2023using}, their use as a foundation for realistic counterfactual benchmark construction remains largely unexplored. Moreover, existing work has primarily focused on evaluating fixed intervention strategies, whereas realistic causal inference settings require reasoning about time-varying and interacting interventions. Our work builds on differentiable ABMs to generate realistic counterfactual trajectories under time-varying and interacting policy interventions.

\begin{figure*}[t]
  \centering
  \captionsetup{labelfont=bf,textfont=normalfont}
  \includegraphics[width=\textwidth]{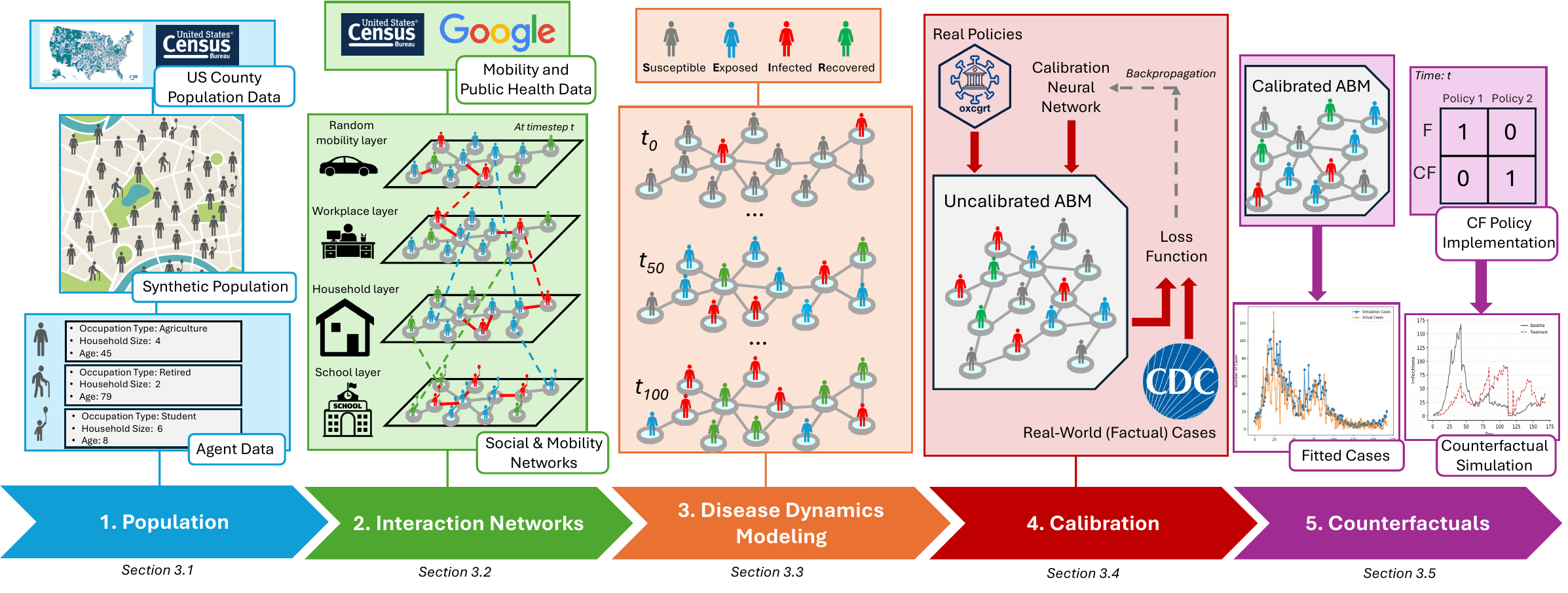}
  \caption{\textbf{Overview of the counterfactual simulation pipeline.} Real-world demographic, mobility, epidemiological, and policy data are integrated into a differentiable agent-based model to construct realistic factual and counterfactual epidemic trajectories. The simulator is calibrated to CDC case data conditioned on the observed policy sequence, and counterfactual scenarios are generated by flipping time-varying intervention policies through corresponding changes in the underlying contact networks.}
  \label{fig:simframework}
\end{figure*}

\subsection{Causal Inference for Time Series}

Estimating the effects of interventions in dynamic systems is a central problem in causal inference, with applications in domains such as healthcare, economics, and epidemiology~\cite{moraffah2021causal, runge2023causal}. A wide range of methods have been proposed to address these challenges, ranging from classical approaches such as marginal structural models, structural nested models, and g-computation~\cite{robins1986new, robins1994correcting, hedeker2006longitudinal, li2021g, frauen2023estimating} to modern machine learning methods based on recurrent neural networks, neural differential equations, Transformers, and generative models~\cite{bica2020estimating, seedat2022continuous, melnychuk2022causal, wu2024counterfactual, mu2025counterfactual}. As a result, the field has progressed from estimating average treatment effects to predicting full counterfactual trajectories and outcome distributions. 

Despite these methodological advances, progress remains constrained by the lack of realistic benchmark datasets with observable counterfactual outcomes~\cite{ferdous2025timegraph}. Existing datasets generally fall into two categories. The first consists of real-world observational datasets, such as IHDP (infant cognitive development)~\cite{brooks1992effects} and MIMIC (electronic health records of ICU patients, including treatment histories and multiple time-varying physiological measurements such as blood pressure and oxygen saturation)~\cite{johnson2016mimic, saeed2002mimic}. These capture complex temporal dynamics but do not provide ground-truth counterfactual outcomes, making direct evaluation of causal predictions impossible. The second consists of synthetic or semi-synthetic datasets, where counterfactuals are available by construction~\cite{wu2024counterfactual, seedat2022continuous, lim2018forecasting, mu2025counterfactual}. However, these benchmarks are often generated from simplified dynamical systems and strong modeling assumptions, limiting their ability to capture realistic intervention effects, confounding structures, and heterogeneous population dynamics. Consequently, existing benchmarks face a fundamental trade-off between realism and evaluability, leaving a gap that our work seeks to address by providing a benchmark with more realistic causal dynamics and observable counterfactual outcomes.

\section{Simulation Framework}\label{sec:method}
Our benchmark \ourmethod is based on an ABM epidemiological simulation that recreates multiple county-level populations, models individual movement and disease spreading dynamics, and generates COVID-19 case trajectories under both factual and counterfactual policy scenarios. Our framework leverages fine-grained agent-based modeling to represent interactions at the agent-to-agent level, enabling principled modeling of COVID-19 infection dynamics, with a tensorized and differentiable design that supports fast and scalable parameter calibration.

As shown in Figure~\ref{fig:simframework}, our counterfactual simulation pipeline begins by generating synthetic populations from real-world population data. We then construct a system of interaction networks, including three static and dynamic social contact networks representing workplace, household, and school interactions, as well as a time-varying random mobility network. These networks define the contact structure over which the disease dynamics model operates, allowing COVID-19 transmission to unfold realistically across a county-wide population. We next calibrate the infectious disease dynamics using a differentiable agent-based model, enabling end-to-end gradient-based fitting to real-world policy and COVID-19 infection data. This calibration yields an ABM that produces realistic population-level transmission dynamics under controlled intervention settings. Building on this calibrated model, we implement policy mechanisms to generate realistic counterfactual scenarios by flipping the factual policy at a given timestep, providing insight into how alternative policies could have affected disease transmission dynamics. Each of the five components is explained in more detail in the following subsections.

\subsection{Synthetic Population Generation} \label{subsec:population}

We generate synthetic populations for each county using demographic attributes derived from the 2020 U.S. Census, including age, household size, and occupation information. For a given county $C$ with population size $N$, we create $N$ agents whose attributes are sampled to match county-level demographic distributions.

Agent ages are sampled according to county-specific age-group proportions, with ages drawn uniformly within each age bracket. Household sizes are assigned based on county-level household size distributions, and agents are grouped into households accordingly, with each household assigned a unique identifier to support the construction of household contact networks. For working-age agents, occupation types are sampled from county-level occupation distributions, with an additional fraction assigned as unemployed to reflect contemporaneous labor statistics. Agents below working age are assigned as students, and older agents are assigned as retired. 

The resulting synthetic population for county $C$ consists of $N$ agents with age, household, and occupation attributes, and serves as the input to the agent-based simulation described in subsequent sections. More details of the sampling procedures and demographic distributions are provided in the appendix.

\subsection{Social Contact and Mobility Networks} \label{subsec:networks}
We construct social contact and mobility networks following a procedure closely aligned with~\cite{hinch2021openabm}. Unlike prior work that relied on detailed networked mobility data, we instead use publicly available time-series mobility data to parameterize population-level movement and interaction patterns. The resulting interaction structure consists of two static social contact networks—household and school—as well as a dynamic social contact network—occupation—and a random, time-varying mobility network, denoted by $\mathcal{G}_H$, $\mathcal{G}_S$, $\mathcal{G}_O$, and $\mathcal{G}_R(t)$, respectively.

\subsubsection{Social contact networks.}
The social networks capture structured, recurrent interactions arising from household membership, school attendance, and occupational activity. Household contacts connect agents within the same household. Occupation and school contact networks are constructed using interaction parameters from~\cite{hinch2021openabm} and age-based U.S. Census data, respectively, using a Watts--Strogatz-style random graph construction to reflect typical contact patterns. What separates the occupation network from the school and household networks, however, is its time-varying nature, where we use Google's mobility data containing the workplace percent change from baseline for a given county. The other two networks (static) remain fixed over time and collectively define the static interaction structure.

\subsubsection{Time-varying mobility network.}
To capture transient and incidental interactions, we construct a random mobility network $\mathcal{G}_R(t)$ at each time step using Google mobility data in combination with Census-derived age-based contact statistics. Baseline interaction rates are modulated over time in response to observed mobility trends, allowing overall contact intensity to vary in line with real-world behavioral changes. Agent-to-agent interactions are sampled stochastically at each time step to form the mobility network.

\subsubsection{Full interaction network.}
At each time step $t \in \{0,\dots,T\}$, the complete interaction network is defined as the union of the static social contact networks and the time-varying mobility network,
\[
\mathcal{G}(t) = \mathcal{G}_{\text{static}} \cup \mathcal{G}_O(t) \cup \mathcal{G}_R(t),
\]
where $\mathcal{G}_{\text{static}} = \mathcal{G}_H \cup \mathcal{G}_S$. The sequence $\{\mathcal{G}(t)\}_{t=0}^{T}$ governs all agent-to-agent interactions driving disease transmission in the simulation. More details are provided in the appendix.

\subsection{Infectious Disease Model}

We simulate infectious disease spread using a differentiable agent-based model that extends the GradABM framework~\cite{chopra2023differentiable}, implemented in AgentTorch~\cite{chopra2024flame}. Disease progression follows a Susceptible–Exposed–Infectious–Recovered (SEIR) compartmental model~\cite{wu2020nowcasting} at the agent level. Each agent occupies one of the SEIR compartments and transitions sequentially through susceptible, exposed, infectious, and recovered states. Initial conditions are specified by a learnable fraction of infected agents, with the remaining population initialized as susceptible. At each time step, transmission occurs through agent-to-agent interactions defined by the time-varying interaction network $\mathcal{G}(t)$, with infection probabilities determined by contact intensity, agent attributes, and network type, following established epidemiological modeling practices~\cite{hinch2021openabm}. Exposed agents become infectious after a fixed latent period, remain infectious for a fixed duration, and then transition to the recovered state.
Transmission dynamics are governed by a time-varying infection rate scale $R_t$, updated weekly. To align simulated infections with reported case counts, we introduce a scalar underreporting factor $k$, such that the observed number of cases at time $t$ is given by $\hat{Y}_t = k N_t$, where $N_t$ denotes the number of newly infected agents. 

The SEIR progression is initialized at $t=0$ with a learnable proportion of infected individuals, $p$. The exact compartment populations are initialized as:
$$S = (1 - p)N \quad \text{and} \quad I = p N$$
At each time step $t$, the probability $P$ that a susceptible agent $a \in S$ transitions to the exposed compartment $E$ following an interaction with an infected agent $b \in I$ on network $n \in \mathcal{G}(t)$ is governed by the transmission rate $\lambda$:
$$P(t, s_i, a_s, n) = 1 - e^{-\lambda(t, s_i, a_s, n)}$$
where $s_i$ indicates the infector's symptom status 
and $a_s$ is the age of the susceptible agent. The rate $\lambda$ is calculated as:
$$\lambda(t, s_i, a_s, n) = \frac{R_t S_{a_{s}} A_{s_i} B_{n}}{\bar{I}} \int_{t-1}^{t} f_{\Gamma}(u; \mu_i, \sigma_i^2) \, du$$
where $\bar{I}$ is the mean number of daily interactions; $f_{\Gamma}(u; \mu,\sigma^2)$ is the probability density function of a gamma distribution; $\mu_i$ and $\sigma_i$ are the mean and width of the infectiousness curve; $R_t$ is a time-varying scale for the overall infection rate; $S_{a_s}$ is the age scale-factor; $A_{s_i}$ is the asymptomatic scale-factor; and $B_n$ is the network-specific scale-factor \cite{hinch2021openabm}. Following exposure, agents remain latent for a fixed duration $D_{\text{E}}$ before remaining infectious for duration $D_{\text{I}}$.

\subsection{ABM Calibration}

We formulate ABM calibration as an optimization problem to estimate the time-varying infection rate scale $R_t$, the underreporting factor $k$, and the initial infection proportion $I_0$. These parameters are learned via gradient-based optimization using a calibration neural network $\phi$, which maps model outputs to observed case counts and enables end-to-end differentiation through the agent-based simulation. During calibration, we condition the simulation on the observed (factual) policy interventions available in the real-world data, while holding the policy implementation fixed; the details of how policies are represented and applied in the simulation are described in the following section.

To account for an initial ``burn-in'' period and ensure stable epidemic dynamics prior to evaluation, simulations are initialized two weeks before the start of the ground-truth observation window. The model is trained over epochs of $T=182$ simulation time steps by minimizing the mean squared error (MSE) between predicted reported cases and observed CDC case counts,
\begin{equation}
    \mathcal{L}(\theta) = \frac{1}{T} \sum_{t=1}^{T} \bigl(\hat{Y}_t - Y_t^{\text{GT}}\bigr)^2,
\end{equation}
where $\hat{Y}_t = k N_t$ denotes the model-predicted reported cases and $Y_t^{\text{GT}}$ denotes the ground-truth daily incidence.

Calibration is made computationally efficient by the tensorized implementation of the agent-based model and the use of automatic differentiation, allowing gradients to be propagated through the full simulation and the calibration network $\phi$~\cite{chopra2023differentiable}. This enables fast, scalable estimation of high-dimensional model parameters using standard gradient descent. State transition parameters governing the exposed and infectious stages ($D_{\text{E}}$ and $D_{\text{I}}$) are selected via a coarse grid search based on values reported in the literature.
Because epidemiological models often face identifiability challenges (where multiple parameter combinations yield similar mathematical fits), we constrained the time-varying infection rate scale ($R_t$) between $0$ and $5$, and the underreporting factor ($k$) between $0.2$ and $1$. These epidemiologically sound bounds prevented the optimizer from explaining surges via extreme reporting anomalies or biologically unrealistic transmission rates.

\subsection{Counterfactual Policy Implementation}

Following calibration, counterfactual scenarios are generated by fixing the learned epidemiological parameters $(R_t, k, I_0)$ from the final training epoch and performing forward simulations under identical initial conditions while intervening only on policy variables. This isolates the causal effect of policy changes on disease transmission dynamics. In contrast to calibration, which uses observed (factual) policies, counterfactual analysis intervenes on policy inputs while holding all other model components fixed.

We consider counterfactual scenarios involving both static and time-varying policies, including school closures and occupation-related interventions. To realistically simulate the effect of these policies on agent interactions, we modify the underlying contact networks via a network-freezing mechanism.

\subsubsection{Policy Representation}

We model policies using data from the Oxford COVID-19 Government Response Tracker, which reports policy intensity levels for school- and workplace-related interventions on a discrete scale from 0 (weakest) to 3 (strongest). To enable direct comparison with baseline methods and simplify intervention analysis, we map these levels to a binary policy indicator.

Let $l \in \{0,1,2,3\}$ denote the reported policy level, and define the binary indicator $A$ at time step $t$ as
\[
A_t =
\begin{cases}
0, & l \in \{0,1\}, \\
1, & l \in \{2,3\},
\end{cases}
\]
where $A_t=0$ represents a weak policy and $A_t=1$ represents a strong policy.

\subsubsection{Network Freezing Mechanism}

Policy interventions are implemented by probabilistically removing edges from the school and occupation contact networks, a process we refer to as \emph{network freezing}. Let
\[
\mathcal{G}_x = (V_x, E_x), \quad x \in \{S, O\},
\]
denote the school ($S$) and occupation ($O$) contact networks, respectively. At each time step $t$, a binary policy indicator $A_t^{(x)} \in \{0,1\}$ determines the strength of the intervention applied to network $x$.

We define a random freezing proportion $F_x(t) \in [0,1]$ as
\[
F_x(t) \sim
\begin{cases}
\mathrm{Uniform}(0, 0.25), & \text{if } A_t^{(x)} = 0, \\
\mathrm{Uniform}(0.75, 1), & \text{if } A_t^{(x)} = 1.
\end{cases}
\]
At time $t$, a subset of edges $R_x(t) \subseteq E_x(t)$ is sampled uniformly at random with
\[
|R_x(t)| = \left\lfloor F_x(t)\,|E_x(t)| \right\rfloor,
\]
and removed to produce the frozen network
\[
\widetilde{\mathcal{G}}_x(t) = (V_x, E_x(t) \setminus R_x(t)).
\]
This stochastic edge removal captures variability in compliance while preserving the underlying network structure.

\subsubsection{Counterfactual Scenario Generation}

Because policies are time-varying and multi-dimensional, our framework generates multiple counterfactual trajectories at each time step. Specifically, for each time step we simulate all combinations of strong and weak school- and occupation-related policies, resulting in three counterfactual scenarios in addition to the observed factual trajectory. This enables systematic evaluation of how different combinations of interventions jointly affect disease transmission dynamics.

\begin{figure*}[htbp]
    \centering
    \includegraphics[width=\textwidth]{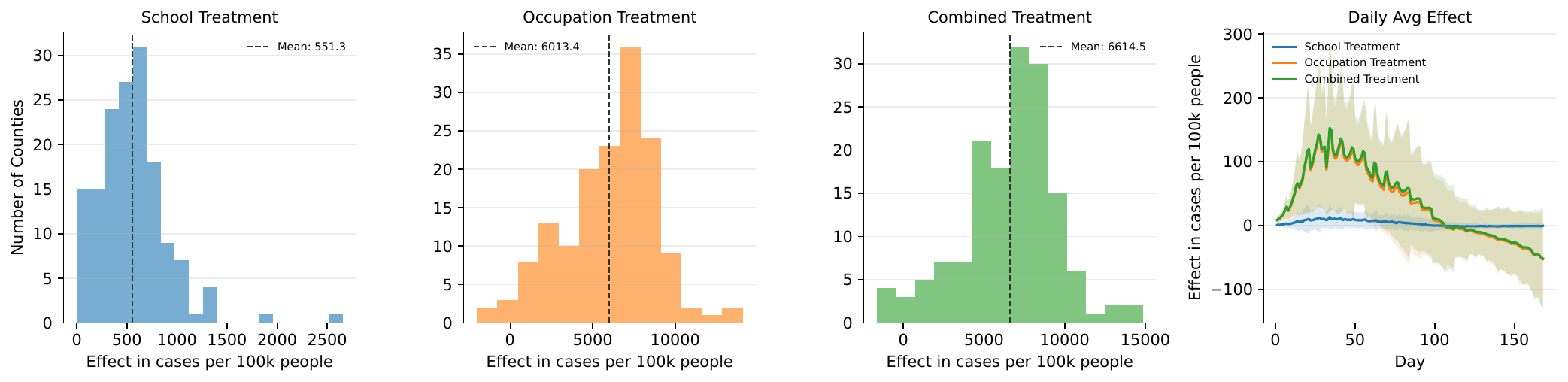}
    \caption{\textbf{Distribution of treatment effects across our 158 U.S. counties.} The first three panels show the density of prevented cases per 100k population for School-only, Occupation-only, and Combined policies. The rightmost panel illustrates the daily average treatment effect trajectory with standard deviation shadows.}
    \label{fig:abm_calibration2}
\end{figure*}

\section{Dataset Description}\label{sec:problem}
One of the main contributions of \ourmethod is the introduction of detailed and realistic counterfactual time-series datasets. Through the previously described construction and calibration of agent-based models grounded in real-world data, we simulate real-world populations using agents whose learned interaction patterns reflect underlying real-world dynamics and causal structures. Based on this framework, we generate factual time-series trajectories that are consistent with observed data under real-world epidemic control policies, and produce corresponding counterfactual time-series trajectories by systematically modifying these policies.

\subsection{Counterfactual Problem Definition}
To facilitate standardized evaluation, we formalize the counterfactual prediction problem considered in \ourmethod. 
\subsubsection{Treatment}
In our time-series setting, both factual and counterfactual treatments are represented as sequences of length T, where each time step has a binary treatment value. Formally: $A_t=\{0,1\},\forall t\in \{0,T\}$. The factual treatment follows the values observed in the real world, while the counterfactual treatment is constructed to highlight its causal effect over the entire time horizon by setting the treatment values at all time steps to either 0 or 1. In addition, depending on whether the problem involves a single policy variable or multiple policy variables, the treatment at each time step can be either one-dimensional or two-dimensional.

\subsubsection{Target Variable}
Both our factual and counterfactual outcome variables are time series of length T, where each time step consists of a one-dimensional integer representing the number of confirmed new COVID-19 cases. Formally, $Y_t\in N,\forall t \in \{0,T\}$. Our analysis shows that the temporal trends of the simulated outcomes under different policy scenarios reflect treatment effects that are consistent with domain expert knowledge. In addition, the agent-based model simultaneously generates other epidemic-related variables, such as the number of exposed individuals and deaths. While these variables could serve as outcome variables for certain high-dimensional causal inference tasks, in this work we treat them as covariates.

\subsubsection{Covariates}
We include covariates to condition causal estimation and control for confounding factors that influence both treatment assignment and outcomes. The dataset contains both static and dynamic covariates. Static covariates are time-invariant features derived primarily from real-world records, including region identifiers, total population size, and age-group distributions. These variables capture persistent demographic and structural characteristics of each county. Dynamic covariates are time-varying features that evolve over the course of the epidemic. These include observed mobility signals as well as epidemic state variables generated by the ABM, such as the numbers of susceptible, exposed, and deceased. Together, these covariates provide the temporal context needed to model treatment effects in a dynamic setting.

\subsection{Single Policy Dataset}
To analyze the isolated causal effect of occupation-based interventions, we constructed a single-policy dataset spanning a 168-day period from October 26, 2020, to April 11, 2021. The geographical scope includes 158 distinct U.S. counties, randomly selected to ensure a diverse representation of states (across 20 U.S. states), with population sizes restricted to the range of 20,000 to 200,000 individuals. To ensure the reliability of our causal predictions, we qualitatively filtered this selection to retain only those counties where the calibrated model demonstrated a high degree of fidelity to ground truth trends, discarding instances where the optimization process failed to converge to a realistic epidemiological baseline.

In this configuration, we control for confounding variables arising from educational interactions by fixing the school policy to the null state for the entire simulation duration. We define the treatment $A_t$ as a one-dimensional binary variable representing the occupation policy at time $t$. With school interactions effectively unconstrained, we generate two distinct epidemiological trajectories for the target variable $Y_t$ (confirmed cases) for each county. First, we simulate the factual scenario, where the treatment sequence $A_t$ follows the time-varying values observed in the real world. Second, we generate the counterfactual scenario by inverting the binary treatment values at each time step. This approach allows us to observe the marginal impact of workplace restrictions on $Y_t$, independent of concurrent changes in school-based mandates.

The resulting dataset exhibits substantial heterogeneity in treatment effects across counties. As shown in Figure~\ref{fig:abm_calibration2}, occupation-based interventions produce a wide range of cumulative effects on COVID-19 incidence, reflecting differences in population structure, mobility patterns, and epidemic trajectories. This variability creates a challenging and realistic setting for evaluating causal inference methods, requiring models to capture both temporal dynamics and heterogeneous treatment responses across populations.

\subsection{Multi Policy Dataset}

Epidemic dynamics are rarely driven by a single variable; therefore, the multi-policy dataset is designed to capture the interaction effects between simultaneous school and occupation interventions. This dataset utilizes the same 158-county selection and 168-observation window defined in the single-policy setting. In this setting, we relax the constraints of the single-policy dataset to evaluate the joint impact of interventions by defining the treatment $A_t$ as a two-dimensional vector comprising both school and occupation policies, $A_t = [A_t^{(S)}, A_t^{(O)}]$.

For every county population, we generate four distinct simulation trajectories representing the full permutation of time-varying policy applications, where the daily school policy $p_S(t)$ and occupation policy $p_O(t)$ are either maintained as their historical sequence or inverted:  
\textbf{Factual-Factual:} Both components of the treatment vector $A_t$ follow real-world historical data sequences.
\textbf{Counterfactual-Factual:} The school policy component $A_t^{(S)}$ is flipped at every time step; the occupation component $A_t^{(O)}$ remains factual.
\textbf{Factual-Counterfactual:} The school policy component $A_t^{(S)}$ remains factual; the occupation component $A_t^{(O)}$ is flipped at every time step.
\textbf{Counterfactual-Counterfactual:} Both components of the treatment vector $A_t$ are flipped for the entire duration.

\begin{figure*}[htbp]
    \centering
    \includegraphics[width=\textwidth]{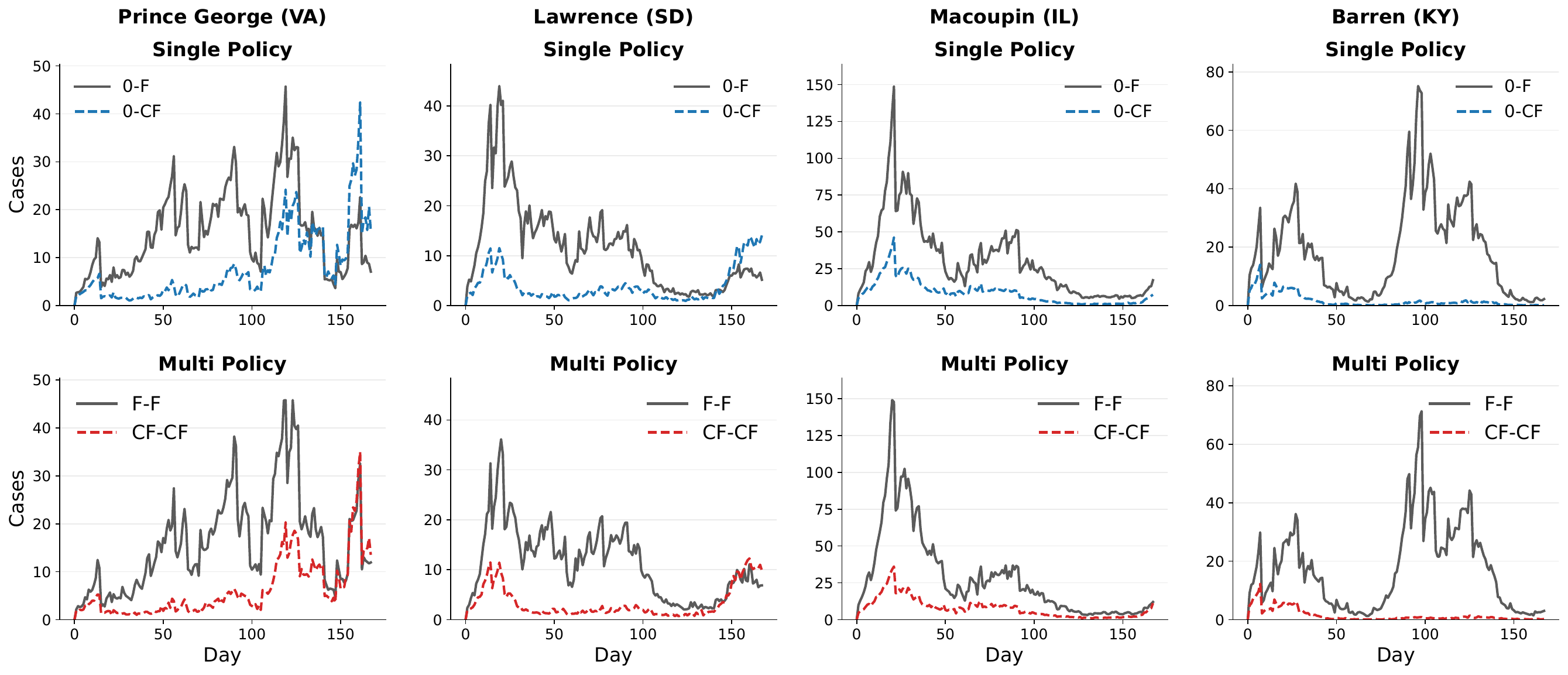}
    \caption{\textbf{Side-by-side comparison of intervention effect in dynamics}. Top Row: Single Policy (Occupation-only) intervention with no school policy. Bottom Row: Multi-Policy (School + Occupation) intervention. \textbf{Notation:} \textit{0-F} denotes open schools with factual occupation closures; \textit{0-CF} denotes open schools with counterfactual (open) occupations. \textit{F-F} denotes the fully factual baseline (historical school and occupation policies); \textit{CF-CF} denotes the fully counterfactual scenario.}
    \label{fig:5}
\end{figure*}

This comprehensive generation process allows for a granular analysis of how simultaneous, time-varying interventions reinforce or diminish one another, providing a robust dataset for training causal inference models on complex, multi-modal intervention strategies. To provide a direct comparison of these intervention effects, Figure~\ref{fig:5} shows the contrast between single-variable and multi-variable scenarios. The top row isolates the impact of occupation-based interventions in a scenario where school policy is held constant (open). The bottom row extends this to our joint multi-policy intervention, comparing the fully factual scenario (\textit{F-F}) against the fully counterfactual scenario (\textit{CF-CF}). Unifying the y-axis across rows allows for a direct visual assessment of how the magnitude of infection suppression scales when moving from a single intervention to combined policies.

\subsection{ABM Calibration Quality} \label{subsec:calibration_quality}

\begin{figure*}[htbp!]
    \centering
    \includegraphics[width=\textwidth]{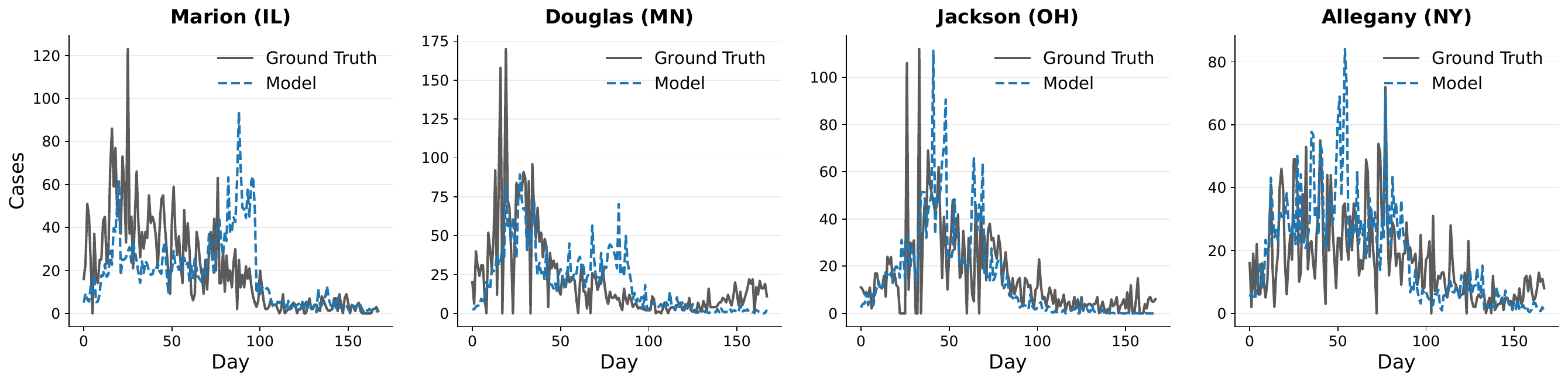}
    \caption{\textbf{Calibration validation for four representative counties.} The plots contrast the ABM's simulated factual trajectory (dashed blue) against the historical ground truth of daily cases (solid black).}
    \label{fig:calibration_grid}
\end{figure*}

Since the realism of \ourmethod depends on the fidelity of the underlying simulator, we evaluate the quality of the calibrated ABMs before analyzing the generated counterfactual datasets. We rigorously validated the calibrated ABM against historical CDC infection data across our 158 selected counties, where the ABM demonstrated robustness across geographies and high accuracy, achieving a mean Normalized Root Mean Square Error (NRMSE, defined as $\text{NRMSE} = \frac{\text{RMSE}}{\bar{y}}$ to allow scale-independent comparison) of 0.1887 with a standard deviation of 0.0685. Beyond quantitative metrics, Figure~\ref{fig:calibration_grid} demonstrates qualitative alignment with actual cases. We ensured a qualitative good fit based on three primary criteria: accurate phase alignment (correct peak infection timing), magnitude consistency (proportional surge amplitudes), and morphological similarity (capturing complex multi-wave behaviors). Importantly, calibration did not succeed for all attempted populations. While standard bell curves presented a smooth optimization landscape, highly irregular multi-wave behaviors were significantly harder to fit. Because the mechanistic nature of an ABM introduces structural rigidity, the optimizer occasionally fell into local minima, generating flat trajectories that missed actual surges. By filtering out these unstable instances, we limited our dataset to the 158 counties that were successfully calibrated. 

As illustrated in Figure~\ref{fig:calibration_grid}, the resulting factual baseline reliably approximates real-world transmission dynamics. Furthermore, our simulations capture critical epidemiological nuances, such as the ``flattening the curve'' effect. In certain scenarios, effective interventions suppressed initial surges but preserved a larger pool of susceptible individuals, occasionally resulting in a negative treatment effect at the tail end of the simulation horizon.

\section{Benchmark and Results}\label{sec:experiments}
\ourmethod is a benchmark suite that consists of two tasks designed to cover common settings in time-series causal inference: (1) causal inference under a single time-varying policy, and (2) causal inference under multiple time-varying policies. We select commonly used and competitive models as baselines and evaluate their performance using a broad set of metrics, establishing a benchmark for future research.
\subsection{Setup}
\noindent\textbf{Problem Formulation.} 
(1) Causal inference under a single time-varying policy: Given static scalar covariates $x_s \in \mathbb{R}^s$, dynamic time-series covariates $x_d \in \mathbb{R}^{d \times T}$, a treatment sequence $A \in \{0,1\}^T$, and the corresponding observed outcome time series $y \in \mathbb{R}^T$, the objective is to predict the outcome trajectory under an alternative treatment sequence $A’ \in \{0,1\}^T$, denoted by $y \mid x_s, x_d, A’$.
(2) Causal inference under multiple time-varying policies: Given static scalar covariates $x_s \in \mathbb{R}^s$, dynamic time-series covariates $x_d \in \mathbb{R}^{d \times T}$, a multivariate treatment sequence $A \in \{0,1\}^{m \times T}$, and the corresponding observed outcome time series $y \in \mathbb{R}^T$, the objective is to predict the outcome trajectory under an alternative treatment sequence $A’ \in \{0,1\}^{m \times T}$, denoted by $y \mid x_s, x_d, A’$.
\\
\noindent\textbf{Baselines.}
We include both classical and state-of-the-art models for time-series causal inference: 
\textit{Kernel Density Estimation (KDE)} is a non-parametric method for estimating an unknown probability density function from observed samples~\cite{davis2011remarks}.
\textit{Transformer} leverages self-attention mechanisms to capture long-range temporal dependencies in time series data~\cite{vaswani2017attention}.
\textit{CVAE} models the potential outcomes distribution by learning a conditional variational approximation~\cite{kingma2013auto}.
\textit{CDiffusion} generates counterfactual outcome by a conditional diffusion process~\cite{ho2020denoising}.
\textit{MSDiffusion} is a generative model that learns causal distributions under time-varying policies via inverse probability of treatment weighting (IPTW)~\cite{wu2024counterfactual}.
\textit{MSVAE} is a conditional VAE model which leverages inverse probability of treatment weighting to realize unbiased learning~\cite{wu2024counterfactual}.
\textit{TE-CDE} uses controlled differential equations to handle irregularly sampled data for counterfactual outcome prediction~\cite{seedat2022continuous}.
\textit{S-learner} uses a single RNN to model the regression of the outcome variable~\cite{kunzel2019metalearners}.
\textit{T-learner} employs two distinct RNN models to separately learn the outcome regressions for the treated and control groups~\cite{kunzel2019metalearners}. RNNs are widely used methods that model temporal dependencies in sequential data by recursively updating hidden states~\cite{sherstinsky2020fundamentals}.

\begin{table}[]
    \centering
    \caption{Comparison of experimental settings applicable to different methods. (\xmark) indicates that the method was not originally developed for this class of problems, but can be adapted to handle them.
    Single, multiple, static, and time-varying represent different types of treatments.
    }
    \begin{tabular}{c|cccc}
    \toprule
    \midrule
       Baselines  & Single & Multi & Static & Time-V \\
       \midrule
       KDE & \cmark & \cmark & \cmark & \cmark\\
       ODE & \cmark & \cmark & \cmark & \cmark\\
       S-learner & \cmark & \cmark & \cmark & \cmark\\
       T-learner & \cmark & \cmark & \cmark & \xmark \\
       RNN & \cmark & \cmark & \cmark & \cmark\\
       Transformer & \cmark & \cmark & \cmark & \cmark\\
       GANITE~\cite{yoon2018ganite} & \cmark & \cmark & \cmark & \xmark\\
       DiffPO~\cite{ma2024diffpo} & \cmark & \xmark & \cmark & \xmark\\
       MS-Diffusion~\cite{wu2024counterfactual} & \cmark & \xmark & (\xmark) & \cmark\\
       MS-VAE~\cite{wu2024counterfactual} & \cmark & \xmark & (\xmark) & \cmark \\
       TE-CDE~\cite{seedat2022continuous} & \cmark & \xmark & (\xmark) & \cmark\\
       R-MSN~\cite{lim2018forecasting}  & \cmark & \xmark & (\xmark) & \cmark\\
       ODE-Diff~\cite{mu2025counterfactual} & \cmark & \xmark & (\xmark) & \cmark\\
       INFs~\cite{melnychuk2023normalizing} & \cmark & \xmark & \cmark & \xmark\\
       \midrule
    \bottomrule
    \end{tabular}
    \vspace{-15pt}
\end{table}

\begin{table*}[]
    \centering

    \begin{minipage}[t]{0.48\textwidth}
    \centering
    
    \caption{Performance of baselines under single time-varying policy. Lower is better, except for 95\% PI.}
    \vspace{-1em}
    \resizebox{\textwidth}{!}{
    \begin{tabular}{c|ccccc}
    \toprule
    \midrule
       Baselines  & WD & RMSE & CATE RMSE & 95\% PI & Calibration Score \\
       \midrule
       KDE & $1.146{\scriptstyle \pm0.000}$ & $1.464{\scriptstyle \pm0.000}$ & $0.790{\scriptstyle \pm0.000}$ & - & -\\
       S-learner & $0.559{\scriptstyle \pm0.044}$ & $0.739{\scriptstyle \pm0.031}$ & $0.999{\scriptstyle \pm0.052}$ & - & - \\
       T-learner & $0.568{\scriptstyle \pm0.016}$ & $0.832{\scriptstyle \pm0.030}$ & $0.917{\scriptstyle \pm0.064}$ & - & - \\
       Transformer & $0.713{\scriptstyle \pm 0.045}$ & $0.817{\scriptstyle \pm 0.043}$  & $0.592{\scriptstyle \pm 0.029}$ & - & - \\
       TE-CDE & $1.814{\scriptstyle \pm0.279}$ & $1.953{\scriptstyle \pm0.283}$ & $0.559{\scriptstyle \pm0.012}$ & - & - \\
       MS-VAE & $1.969{\scriptstyle \pm0.123}$ & $2.143{\scriptstyle \pm0.138}$ & $0.665{\scriptstyle \pm0.041}$ & $53.7\%{\scriptstyle \pm2.8}$ & $0.285{\scriptstyle \pm0.014}$ \\
       MS-Diffusion & $1.007{\scriptstyle \pm0.250}$ & $1.215{\scriptstyle \pm0.252}$ & $0.637{\scriptstyle \pm0.147}$ & $34.6\%{\scriptstyle \pm7.0}$ & $0.407{\scriptstyle \pm0.040}$ \\
       \midrule
    \bottomrule
    \end{tabular}
    \label{fig:results_single}
    }
    \end{minipage}
    \hfill
    \begin{minipage}[t]{0.48\textwidth}
    \centering
    \caption{Performance of baselines under multiple time-varying policies. Lower is better, except for 95\% PI.}
    \vspace{-1em}
    \resizebox{\textwidth}{!}{
    \begin{tabular}{c|ccccc}
    \toprule
    \midrule
       Baselines  & WD & RMSE & CATE RMSE & 95\% PI & Calibration Score\\
       \midrule
       KDE & $0.922{\scriptstyle \pm0.000}$ & $1.167{\scriptstyle \pm0.000}$ & $0.666{\scriptstyle \pm0.000}$ & - & -\\
       S-learner & $0.594{\scriptstyle \pm0.035}$ & $0.734{\scriptstyle \pm0.037}$ & $0.683{\scriptstyle \pm0.011}$ & - & - \\
       T-learner & $0.873{\scriptstyle \pm0.059}$ & $1.028{\scriptstyle \pm0.047}$ & $0.926{\scriptstyle \pm0.017}$ & - & - \\
       Transformer & $0.902{\scriptstyle \pm0.160}$ & $1.048{\scriptstyle \pm 0.129}$ & $0.768{\scriptstyle \pm 0.111}$ & - & -\\
       CVAE & $1.811{\scriptstyle \pm0.041}$ & $2.891{\scriptstyle \pm 0.051}$ & $2.226{\scriptstyle \pm0.057}$ & $66.8\%{\scriptstyle \pm10.8}$ & $0.241{\scriptstyle \pm0.040}$ \\
       CDiffusion & $1.257{\scriptstyle \pm0.141}$ & $1.509{\scriptstyle \pm0.118}$ & $1.128{\scriptstyle \pm0.074}$ & $34.5\%{\scriptstyle \pm6.6}$ & $0.405{\scriptstyle \pm0.036}$ \\
       \midrule
    \bottomrule
    \end{tabular}
    \label{fig:results_multi}
    }
    \end{minipage}
\end{table*}

\noindent\textbf{Metrics.} Across different benchmarks, we adopt a broad and standard set of evaluation metrics to assess the performance of different models on causal inference tasks, including Wasserstein Distance (WD), 
root mean squared error (RMSE), predictive interval coverage (confidence level of 95\%), calibration scores, and RMSE of Conditional Average Treatment Effects (CATE). All results were normalized prior to evaluation to provide a reasonable reference.

RMSE and $k$-WD both measure the discrepancy between model predictions and the benchmark values. In our experiments, we report the Wasserstein Distance with $k=1$, which is a common choice in the literature~\cite{ma2024diffpo}. For both RMSE and WD, smaller values indicate better performance. CATE is a standard target in causal inference and is computed according to its definition. Benefiting from the availability of counterfactual outcomes in our dataset, we are able to compute the RMSE between the predicted CATE and the ground-truth CATE, where smaller values indicate stronger causal inference capability. For methods that are capable of generating predictive distributions, we use predictive interval (PI) coverage to evaluate their uncertainty quantification performance in causal inference tasks. In addition, the calibration score further assesses how well the predicted intervals reflect uncertainty across different confidence levels. Following prior work, we consider 11 different confidence levels in our evaluation~\cite{li2025neural}. All metrics are reported with mean and standard deviation over three runs.

\subsection{Single-Policy Time-Varying Results}
Table~\ref{fig:results_single} presents the performance of different models under a single time-varying policy. 
We observe that the proposed dataset captures complex temporal dynamics and causal relationships, making it difficult for simple methods such as KDE to adequately model the underlying data-generating process. 
On the other hand, RNN-based meta-learners and Transformer achieve strong performance on the WD and RMSE metrics, which we attribute to their ability to model complex temporal dynamics. The meta-learners achieve relatively strong predictive accuracy, but their higher CATE error is mainly due to large errors at a small number of counties and time points, which drag down the overall performance. 
Regarding TE-CDE, although it was designed for counterfactual prediction in temporal settings, its lower predictive accuracy suggests that its representational capacity for highly complex and volatile data remains weaker than that of RNNs and Transformers. However, TE-CDE achieves the best performance in CATE estimation, demonstrating its strength as a causal inference model and further validating that our dataset captures meaningful and complex real-world causal relationships.
Turning to generative models, MS-VAE exhibits relatively weak predictive accuracy, although its wider prediction intervals result in slightly better PI coverage than other generative approaches such as MS-Diffusion. In contrast, our implementation of MS-Diffusion leverages a backbone specifically designed for time-series generation, leading to improved predictive accuracy and CATE estimation performance. However, its prediction interval coverage remains limited, highlighting the difficulty of uncertainty quantification in complex temporal settings.

The experimental results suggest that \textsc{EpiCF-Bench} captures both meaningful causal relationships and realistic temporal dynamics. As a result, strong predictive performance does not necessarily translate into strong causal inference performance, highlighting the need for methods that effectively balance both objectives in complex time-series settings.

\subsection{Multi-Policy Time-Varying Results}
Table~\ref{fig:results_multi} reports the performance of different models under the multi-policy time-varying intervention setting. 

In contrast to the single-policy benchmark, we observe that while the S-learner maintains relatively stable performance, the T-learner experiences a substantial degradation across most metrics. One possible explanation is that learning separate outcome functions for different treatment regimes reduces the effective sample size available to each model, making it more difficult to capture the complex temporal dynamics and treatment interactions present in the multi-policy setting. 
More broadly, several methods, including Transformer-, diffusion-, and VAE-based approaches, exhibit noticeable performance degradation under multiple time-varying interventions. As expected, treatment interactions and increased distributional heterogeneity make counterfactual prediction and treatment effect estimation more difficult. \textsc{EpiCF-Bench} therefore provides a valuable testbed for developing and evaluating methods capable of reasoning under multiple interacting interventions.

\section{Conclusion and Discussion}\label{sec:conclusion}

We introduced a principled benchmark for counterfactual prediction in epidemic time series under time-varying interventions, grounded in a calibrated, differentiable agent-based model. By integrating real-world demographic data, mobility signals, and policy interventions, our framework generates high-fidelity factual and counterfactual trajectories that embed realistic causal mechanisms at the agent level. This approach addresses key limitations of existing synthetic and semi-synthetic benchmarks, which often rely on simplified dynamics or estimated counterfactuals that fail to capture the complexity of real-world epidemic processes. 

Using the resulting dataset, we conducted a comprehensive evaluation of widely used and state-of-the-art time-series causal inference methods across single- and multi-policy settings. Our results reveal substantial performance differences among methods and highlight the importance of realistic benchmarks for evaluating counterfactual prediction under dynamic and interacting interventions. Beyond benchmarking, the proposed pipeline provides a reusable and extensible foundation for generating causal time-series data that preserves mechanistic structure while remaining amenable to learning-based analysis.

Looking ahead, the proposed framework naturally extends to other infectious diseases and epidemiological settings. Recent efforts to provide richer mobility and behavioral dynamics data~\cite{gozzi2025epydemix}, as well as parallel initiatives focused on collecting and curating detailed policy and intervention information~\cite{howerton2023evaluation}, open new opportunities for constructing even more realistic and diverse counterfactual benchmarks. More broadly, this work points toward tighter integration between mechanistic simulators and data-driven causal models, enabling more reliable evaluation and development of methods for decision-making in complex, real-world dynamical systems.

\section{Limitations and Ethical Considerations}
\paragraph{Limitations}
Despite its realism, our framework necessarily relies on modeling assumptions and data abstractions. While the ABM is calibrated to observed case data and grounded in established epidemiological practices, it cannot capture all sources of heterogeneity present in real populations, such as fine-grained behavioral adaptation, informal social interactions, or unobserved policy compliance dynamics. In addition, policy interventions are implemented via stochastic edge removal in contact networks, which approximates changes in interaction intensity but does not explicitly model individual-level behavioral responses or enforcement mechanisms. As with any simulation-based benchmark, conclusions drawn from this dataset should be interpreted as reflecting the modeled causal structure rather than ground-truth real-world effects. In addition, due to the computational cost of training on large-population counties and running the agent-based model, we set the upper bound of county population to 200,000. Expanding the scale of the dataset could be part of our future work.

\paragraph{Ethical considerations}
This work uses only aggregated, publicly available data sources, including census demographics, mobility trends, and policy indicators, and does not involve any personally identifiable information. The generated synthetic populations are not intended to represent specific individuals or communities. We emphasize that the benchmark is designed for method evaluation and scientific analysis, not for direct policy prescription. Future extensions should continue to consider fairness, transparency, and potential downstream impacts when applying causal inference models to sensitive public health decision-making settings.

\begin{acks}
This publication was made possible by the Insight Net cooperative agreements NU38FT000002 from the CDC’s Center for Forecasting and Outbreak Analytics (CDC-RFA-FT-23-0069). Its contents are solely the responsibility of the authors and do not necessarily represent the official views of the Centers for Disease Control and Prevention. This work was also supported in part by the Summer Undergraduate Research Experience (SURE) program at the University of Michigan. We also acknowledge the AgentTorch developers for creating and maintaining the open-source repository upon which parts of our codebase build.
\end{acks}


\newpage
\bibliographystyle{ACM-Reference-Format}
\bibliography{reference}


\section*{Appendix A: Temporal Dynamics of Negative Treatment Effects}

This appendix presents county-level infection trajectories for specific instances where the cumulative treatment effect was observed to be negative. These cases illustrate the epidemiological trade-off described in Section~\ref{subsec:calibration_quality}, where the intervention successfully "flattens the curve" rather than simply reducing total volume.

In the figures below, the baseline "No Policy" scenario (solid blue line) exhibits a sharp, early peak followed by a rapid decline as the population reaches saturation. In contrast, the "Combined Treatment" scenario (dashed red line) suppresses the initial surge, preserving a larger pool of susceptible individuals. This delay allows the epidemic to persist for a longer duration, occasionally resulting in a higher total cumulative count despite successfully mitigating the immediate strain on healthcare capacity.

\begin{figure}[htbp]
    \centering
    \includegraphics[width=\linewidth]{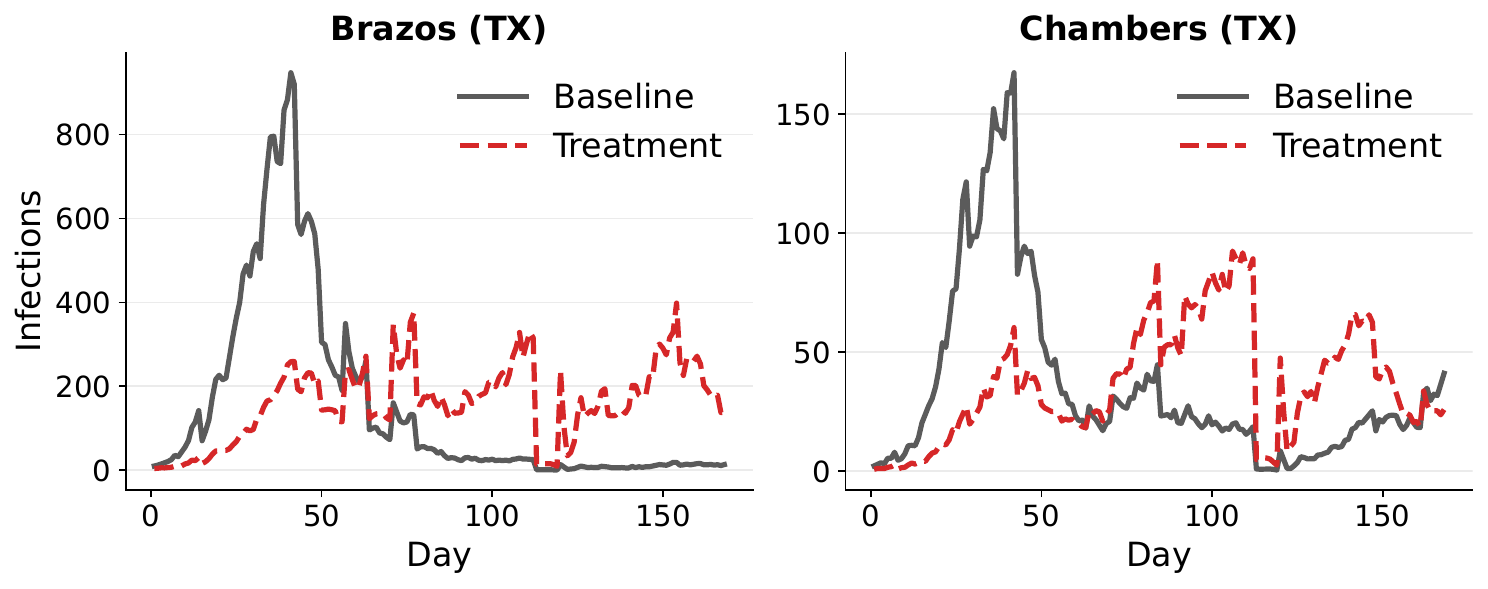}
    \caption{Comparison of "No Policy" vs. "Combined Treatment" scenarios for two counties where intervention policies effectively delay and reduce peak infections while resulting in a larger total volume of cases over the simulation.}
    \label{negative_counties}
\end{figure}

\section*{Appendix B: More Details on Simulation Framework}

\subsection*{B.1 Synthetic Population Generation Formalisms}

Building upon the demographic integration described in Section~\ref{subsec:population}, the synthetic population of size $N$ for county $C$ is mathematically parameterized as follows. 

Age distributions are governed by $A_C$, representing the proportions of the population in 10-year age brackets:
$$A_C = \left\{ \begin{array}{l} 0-9: \pi^A_1 \\ 10-19: \pi^A_2 \\ \vdots \\ 80+: \pi^A_9 \end{array} \right.$$
where $\pi^A_k$ denotes the proportion of the population in age group $k$. The exact age of agent $i$ is assigned via uniform sampling within their selected bracket:
$$\text{age}_i = \text{Uniform}(L_x, U_x), \quad x \in \{0-9, 10-19, \dots, 80+\}$$

Household size assignment relies on proportions $H_C$:
$$H_C = \left\{ \begin{array}{l} 1: \pi^H_1 \\ 2: \pi^H_2 \\ \vdots \\ 6+: \pi^H_6 \end{array} \right.$$
The discrete number of agents assigned to each household size $k$ is calculated as:
$$\text{household\_size}_k = \left\lceil N \cdot H_C(k) \right\rceil, \quad k \in \{1, 2, \dots, 6\}$$

For agents of working age (20--64), occupation types are assigned based on proportions $O_C$, covering 18 specific industries (Agriculture, Mining, Utilities, Construction, Manufacturing, Wholesale trade, Retail/trade, Transportation, Information, Finance/insurance, Real estate/rental, Scientific/technical, Enterprise/management, Waste management, Education, Healthcare, Art, Food) and an ``Other'' category. To align with national data from December 2020, 7\% of this working population pool is reassigned as unemployed. Agents aged 0--19 are strictly designated as ``Student'', and those 65+ as ``Retired''.

\subsection*{B.2 Network Interaction Parameters}

The static and dynamic networks detailed in Section~\ref{subsec:networks} are constructed using specific graph-theoretic constraints. The static graph $\mathcal{G}_{\text{static}}$ is formulated as a simple graph without multiple edges:
$$V_{\text{static}} = \bigcup_{x \in \{H,S\}} V_x, \qquad E_{\text{static}} = \bigcup_{x \in \{H,S\}} E_x$$
The Watts--Strogatz constructions for the occupation ($\mathcal{G}_O$) and school ($\mathcal{G}_S$) networks rely on baseline interaction parameters (average degree $\mu$ and variance $\sigma$) derived from U.S. Census age-based contact statistics and rewiring probabilities established in \cite{hinch2021openabm}. For the random mobility network $\mathcal{G}_R(t)$, these baseline parameters $\mu$ and $\sigma$ are dynamically scaled at each time step $t$ using the Google mobility percent-change-from-baseline signals to appropriately modulate interaction intensity.

\end{document}